\documentclass[10pt,twocolumn,letterpaper]{article}

\usepackage{cvpr}
\usepackage{times}
\usepackage{epsfig}
\usepackage{graphicx}
\usepackage{amsmath}
\usepackage{amssymb}
\usepackage{enumitem}
\usepackage{url}

\usepackage{arydshln}
\usepackage{multirow}

\usepackage[pagebackref=true,breaklinks=true,letterpaper=true,colorlinks,bookmarks=false]{hyperref}

\cvprfinalcopy 

\def\cvprPaperID{1749} 

\ifcvprfinal\fi
\begin{document}

\title{Generation and Comprehension of Unambiguous Object Descriptions}

\author{
Junhua Mao$^2$\protect\footnotemark\ \ \ Jonathan Huang$^1$\ \ Alexander Toshev$^{1}$\ \ Oana Camburu$^{3}$\ \ Alan Yuille$^{2,4}$\ \ Kevin Murphy$^1$\\
{\fontsize{11}{13}\selectfont$^1$Google Inc.\ \ \ \ $^2$University of California, Los Angeles\ \ \ \ $^3$University of Oxford\ \ \ \ $^4$Johns Hopkins University}\\
{\fontsize{10}{11}\selectfont \texttt{\{mjhustc@,yuille@stat.\}ucla.edu}, \texttt{oana-maria.camburu@cs.ox.ac.uk}}\\
{\fontsize{10}{11}\selectfont \texttt{\{jonathanhuang,toshev,kpmurphy\}@google.com}}
}

\maketitle

\newcommand{\eat}[1]{}
\newcommand{\argmax}{\mathrm{argmax}}

\newcommand{\TODO}[1]{\textcolor{red}{\textbf{TODO}:~#1}}
\newcommand{\KEVIN}[1]{\textcolor{blue}{\textbf{Kevin comment}:~#1}}
\newcommand{\ALEX}[1]{\textcolor{blue}{\textbf{Alex comment}:~#1}}
\newcommand{\JONATHAN}[1]{\textcolor{blue}{\textbf{JONATHAN comment}:~#1}}
\newcommand{\VIGNESH}[1]{\textcolor{blue}{\textbf{Vignesh      comment}:~#1}}

\begin{abstract}
We propose a method that can generate an unambiguous description
(known as a referring expression)
of a specific object or region in an image,
and which can also comprehend or interpret such an expression to infer
which object is being described.
We show that our method outperforms previous methods that
generate descriptions of objects without taking into account other
potentially ambiguous objects in the scene.
Our model is inspired by recent successes of deep learning methods for image captioning, 
but while image captioning is difficult to evaluate,  our task allows
for easy objective evaluation.
We also present a new large-scale dataset for referring expressions, based on
MS-COCO.
We have released the dataset and a toolbox for visualization and evaluation, see \url{https://github.com/mjhucla/Google_Refexp_toolbox}.
\end{abstract}

\vspace{-0.5cm}

{\let\thefootnote\relax\footnotetext{The major part of this work was done while J. Mao and O. Camburu were interns at Google Inc.}}

\section{Introduction}\label{sec:intro}

There has been a lot of recent interest in generating text
descriptions of images (see e.g.,
\cite{farhadi2010every,vinyals2014show,donahue2014long,chen2014learning,
  fang2014captions,karpathy2014deep,kiros2014unifying,mao2014deep,xu2015show,devlin2015exploring}).  
However, fundamentally this problem of image captioning is subjective and
 ill-posed.  With so many valid ways to describe any given image, automatic captioning methods are thus notoriously
difficult to evaluate. In particular, how 
can we decide that one sentence is a better description of an image
than another? 

 
In this paper, we  focus on a special case of text
generation given images, where the goal is to generate an unambiguous text
description that applies to exactly one object or region in the image.
Such a description is known as a ``referring expression''
\cite{van2006building,viethen2008use,mitchell2010natural,mitchell2013generating,fitzgerald2013learning,golland2010game,kazemzadeh2014referitgame}.
This approach has a major advantage over generic image captioning,
since there is a well-defined performance metric:
a referring expression is considered to be good if it uniquely
describes the relevant object or region within its context,
such that a listener can comprehend the description and then recover
the location of the original object.
In addition,
because of the discriminative nature of the task, referring
expressions tend to be more detailed (and therefore more useful) than
image captions.
Finally, it is easier to  collect training data to ``cover'' the space of reasonable
referring expressions for a given object than it is for a whole image.

\begin{figure}[t]
\begin{center}
\includegraphics[width=0.95\linewidth]{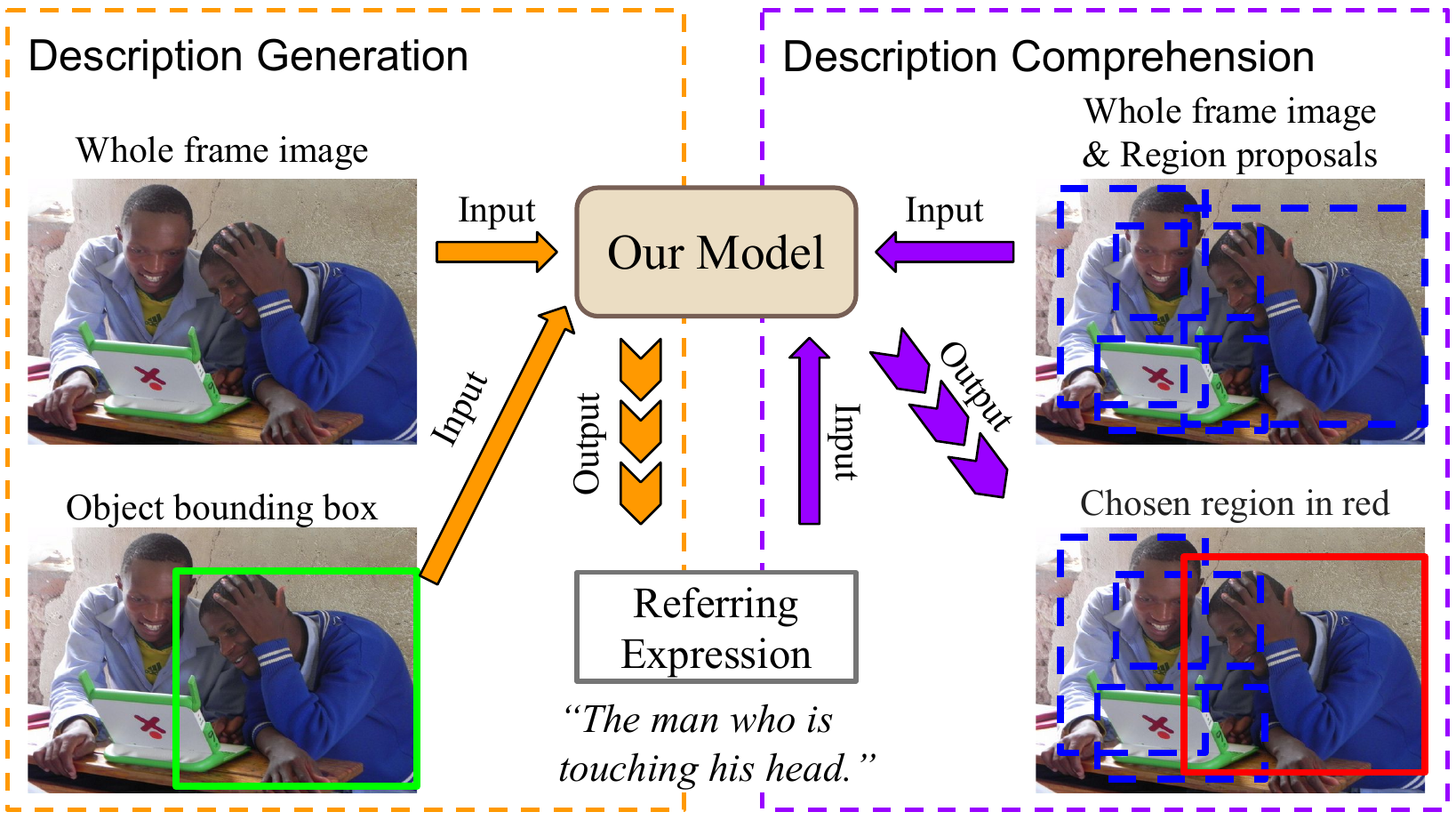}
\end{center}
\vspace{-0.3cm}
   \caption{Illustration of our generation and comprehension system.
On the left we see that the  system is given an image and a region of
interest; it describes it as ``the man who is touching his
head'', which is unambiguous (unlike other possible expressions, such
as ``the man wearing blue'', which would be unclear).
On the right we see that the system is given an image, an expression,
and a set of candidate regions (bounding boxes), and it selects the
region that corresponds to the expression.
}
\vspace{-0.5cm}
\label{fig:intro}
\end{figure}

\eat{
This more specific setting of referring expressions makes evaluation much easier
for several reasons.
First, the space of reasonable
referring expressions of an object tends to be much smaller than the space of reasonable (full) image captions, making it feasible to achieve high
``coverage'' of reasonable ground truth expressions via crowdsourcing.
Second, generating a referring expression requires a speaker to describe the
features of an object that discriminate it from other objects in the scene.
Consequently, the generated descriptions tend to be more detailed than
standard image captions.
Finally, as was exploited by~\cite{kazemzadeh2014referitgame}, referring expressions can be validated objectively by verifying whether it is possible for a listener to identify the referred object within an image.
}
 
We consider two problems: (1) \emph{description
generation}, in which we must generate a text expression that uniquely
pinpoints a highlighted object/region in the image and (2)
\emph{description comprehension}, in which we must automatically
select an object given a text expression that refers to this object
 (see
Figure~\ref{fig:intro}). 
Most prior work in the literature has focused exclusively on 
description generation (e.g.,
\cite{Krahmer2012,kazemzadeh2014referitgame}).  Golland \etal
~\cite{golland2010game} consider  generation and comprehension, but they do not
process real world images.

In this paper, we jointly model both tasks of description generation
and comprehension, using state-of-the-art deep learning approaches 
to handle real images and text. Specifically, our model is based upon
recently developed methods that combine
convolutional neural networks (CNNs) with
recurrent neural networks (RNNs).
We  demonstrate that our model outperforms a baseline which
generates referring expressions without regard to the listener
who must comprehend the expression.
We also show that our model can be trained in a semi-supervised
fashion, 
by automatically generating descriptions for image regions.

Being able to generate and comprehend object descriptions is
critical in a number of applications that use natural language
interfaces, such as controlling a robot (e.g., ``Rosie, please fetch
me the beer from the top shelf of the fridge'',
cf. \cite{Barrett2015}), or interacting with photo editing software
(e.g., ``Picasa, please replace the third car behind the fence with a
motorbike'',  cf. \cite{Cheng2014}).
In addition, it is a good test bed for performing research in the area
of vision and language systems because of the existence of a
useful objective performance measure.

\eat{
It is advantageous to jointly model both generation and comprehension
tasks, and our approach can be contrasted against speaker-only models
that generate but do not comprehend, and thus might generate excellent
descriptions of an object that are not sufficient for discriminating
the object from other objects in the same image.
}
In order to train and evaluate our system, we have collected
and released a new large
scale referring expressions dataset based on the
popular MS-COCO dataset~\cite{lin2014microsoft}.

To summarize, our main contributions are as follows.
First, we present a new large scale dataset for referring expressions.
Second, we evaluate how existing image captioning methods perform at
the referring expression task.
Third, we develop a new method for joint 
generation and comprehension that outperforms current  methods.

\section{Related Work}

\noindent \textbf{Referring expressions.}  Referring expression generation is
a classic NLP problem (see e.g.,
\cite{winograd1972understanding,Krahmer2012}).
Important issues include understanding what
types of attributes people typically use to describe visual objects
(such as color and size)~\cite{mitchell2013generating}, usage of
higher-order relationships (e.g., spatial
comparison)~\cite{viethen2008use}, and the phenomena of over and
under-specification, which is also related to speaker
variance~\cite{fitzgerald2013learning}.

Context (sometimes called pragmatics~\cite{goodman2014probabilistic})
plays a critical role in several ways~\cite{krahmer2002efficient}.
First, the speaker must differentiate the target object from a
collection of alternatives and must thus reason about how the object
differs from its context.  Second, the perception of the listener is
also valuable.  In particular, Golland \etal~\cite{golland2010game}
recently proposed a game theoretic formulation of the referring
expression problem showing that speakers that act optimally with
respect to an explicit listener model naturally adhere to the Gricean Maxims
of communication~\cite{grice1970logic}.

In most of this previous work, authors have focused on small datasets
of computer generated objects (or photographs of simple objects)
~\cite{van2006building,mitchell2010natural} and have not connected
their text generation systems to real vision systems.  However there
has been recent interest in understanding referring expressions in the
context of complex real world images, for which humans tend to
generate longer phrases \cite{Gkatzia2015}. 
Kazemzadeh \etal~\cite{kazemzadeh2014referitgame} were the first to collect a
large scale dataset of referring expressions for complex real world
photos.

We likewise collect and evaluate against a large scale dataset. 
However we go beyond expression generation and jointly learn
both generation and comprehension models.  And where prior works have
had to explicitly enumerate attribute categories such as size, color 
(e.g.~\cite{sadovnik2012image}) or manually list all possible visual 
phrases (e.g.~\cite{sadeghi2011recognition}), 
our deep learning-based models are able to learn to directly
generate surface expressions from raw images without having to first
convert to a formal object/attribute representation.

Concurrently, \cite{hu2015natural} propose a CNN-RNN based method that
is similar to our baseline model and achieve state-of-the-art results
on the ReferIt dataset \cite{kazemzadeh2014referitgame}.
But they did not use the discriminative training strategy proposed in our full model. 
\cite{johnson2015densecap,krishna2016visual} investigate
the task of generating dense descriptions in an image.
But their descriptions are not required to be unambiguous.

\noindent \textbf{Image captioning.} Our methods are inspired by a long line
of inquiry in joint models of images and text, primarily in the vision
and learning
communities~\cite{farhadi2010every,hodosh2013framing,socher2014grounded,Ordonez:2011:im2text,kulkarni2011baby,yang2011corpus,li2011composing}.
From a modeling perspective, our approach is closest to recent works
applying RNNs and CNNs to this problem
domain~\cite{vinyals2014show,donahue2014long,chen2014learning,fang2014captions,karpathy2014deep,kiros2014unifying,mao2014deep,xu2015show}.
The main approach in these papers is to
represent the image content using the hidden activations of a CNN, and then to feed this as input to an RNN, which is trained to generate a sequence of words.

Most papers on image captioning have focused on describing the full image, without any spatial localization.
However, we are aware of two exceptions.
~\cite{xu2015show} propose an attention model
which is able to associate words  to spatial regions within
an image; however, they still focus on the full image captioning task.
~\cite{karpathy2014deep} propose a
model for aligning words and short phrases within sentences to
bounding boxes; they then train an model to generate these
short snippets given features of the bounding box.  
Their model is similar to our baseline model, described in Section~\ref{sec:baseline}
(except we provide the alignment of phrases to boxes in the training set,
similar to \cite{Plummer2015}).
However, we show that this approach is not as good as our full model,
which takes into account other potentially confusing regions in the image.

\noindent \textbf{Visual question answering.} Referring expressions is related to the task of VQA
(see e.g., \cite{antol2015vqa,malinowski2014multi,malinowski2015ask,geman2015visual,gao2015you}).
In particular, referring expression comprehension
can be turned into a VQA task
where the speaker asks a question such as ``where in the image is the car in red?''
and the system must return a bounding box (so the answer is numerical, not linguistic).
However there are philosophical and practical 
differences between the two tasks.  A referring
expression (and language in general) is about \emph{communication} --- in our problem, the
speaker is finding the optimal way to communicate to the listener,
whereas VQA work typically focuses only on answering questions without regard to the listener's state of mind.
Additionally, since questions tend to be more open ended in VQA, evaluating their answers can be as hard as with general image captioning, whereas evaluating the accuracy of a bounding box is easy.

\begin{figure*}[t]
\begin{center}
\includegraphics[width=0.98\linewidth]{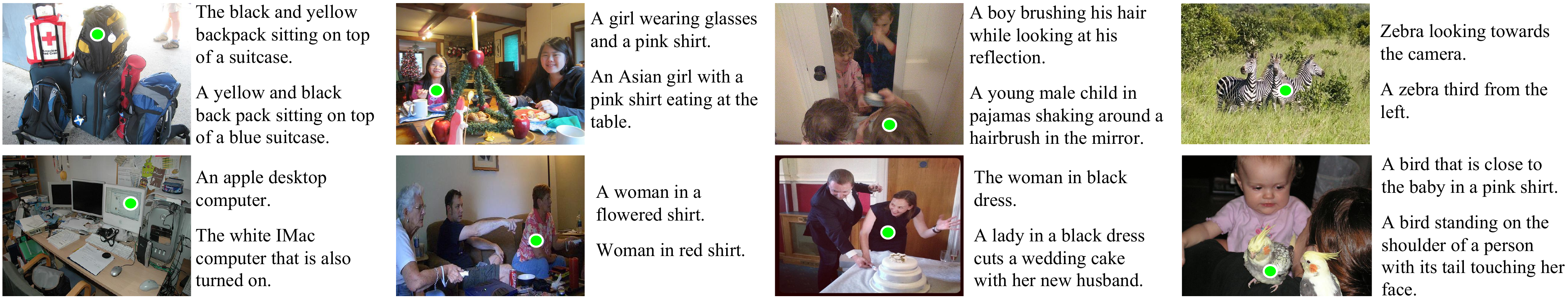}
\end{center}
\vspace{-0.3cm}
 \caption{Some sample images from our Google Refexp (G-Ref) dataset.
          We use a green dot to indicate the object that the descriptions refer to.
          Since the dataset is based on MS COCO, we have access to
the original annotations such as the object mask and category.
          Some of the objects are hard to describe,
          e.g., in the third image in the first row, we need to distinguish the boy from his reflection in the mirror.
}
\vspace{-0.4cm}
\label{fig:dataset_ours}
\end{figure*}

\section{Dataset Construction}
\label{sec:datasets}

\begin{figure}[t]
\begin{center}
\includegraphics[width=0.98\linewidth]{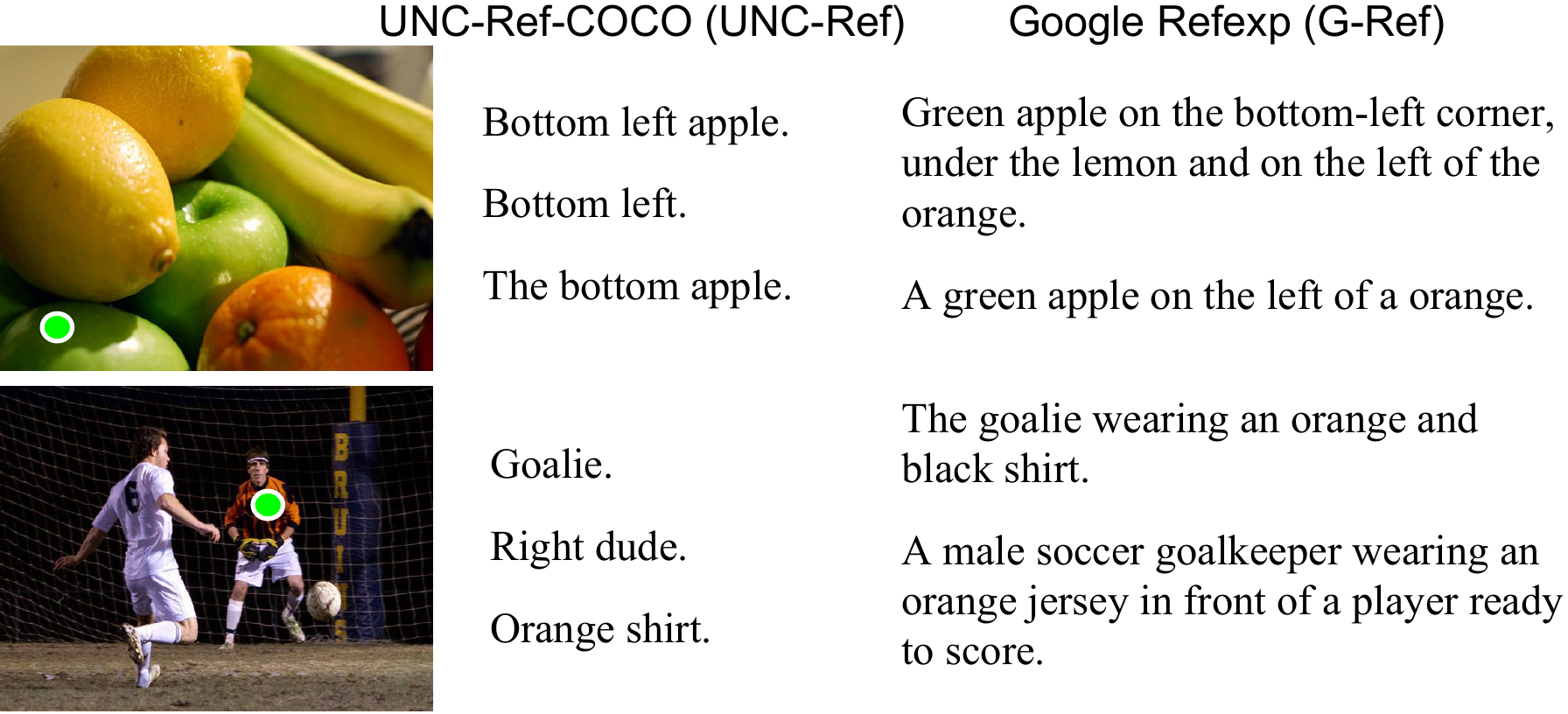}
\end{center}
\vspace{-0.3cm}
 \caption{Comparison between the G-Ref and UNC-Ref dataset.
}
\vspace{-0.5cm}
\label{fig:dataset_comp}
\end{figure}

The largest existing referring expressions dataset that we know of is the \emph{ReferIt dataset},
which was collected by~\cite{kazemzadeh2014referitgame},
and contains 130,525 expressions, referring
to 96,654 distinct objects, in 19,894 photographs
of natural scenes. Images in this dataset are  from 
the segmented and annotated TC-12 expansion of the ImageCLEF IAPR dataset
\cite{Escalante2010}.
Two drawbacks of this dataset, however, are that (1) 
 the images sometimes only contain one object of a given class, allowing speakers to use 
short descriptions without risking ambiguity, and (2)
the ImageCLEF dataset focuses mostly on ``stuff'' (i.e. context) rather than ``things'' (i.e. objects).

In this paper, we use a similar methodology to that of~\cite{kazemzadeh2014referitgame},
but building instead on top of the MSCOCO dataset
\cite{lin2014microsoft},
which contains more than 300,000 images, with 80 categories of objects
segmented at the instance level.

For each image, 
we selected objects if (1) there are between 2 and 4 instances of the same
object type within the same image, and (2) if their bounding boxes occupy at
least 5\% of image area.
This resulted in selecting 54,822 objects from 
26,711 images.
We constructed a Mechanical Turk task in which we presented each
object in each image (by highlighting the object mask) to a worker whose task was to generate a unique text
description of this object.
We then used a second task in which a different worker was presented
with the image and description, and was asked to click inside the
object being referred to. If the selected point was inside the original object's segmentation mask, we considered the
description as valid, and kept it, otherwise we discarded it and re-annotated it by another worker.
We repeated these description generation and verification tasks on
Mechanical Turk iteratively up to three times.
In this way, we selected 104,560 expressions.
Each object has on average 1.91 expressions, and each image has on
average 3.91 expressions.
This dataset is denoted as Google Refexp dataset and some samples are shown in Figure \ref{fig:dataset_ours}.
We have released this dataset and a toolbox for visualization and evaluation, see
\url{https://github.com/mjhucla/Google_Refexp_toolbox}.

While we were collecting our dataset,
we learned that Tamara Berg had independently applied her ReferIt game \cite{kazemzadeh2014referitgame} to the MSCOCO
dataset to generate expressions for 50,000 objects from 19,994 images.
She kindly shared her data (named as UNC-Ref-COCO dataset) with us.
For brevity, we call our Google Refexp dataset as \textit{G-Ref} and the UNC-Ref-COCO as \textit{UNC-ref}.
We report results on both datasets in this paper.
However,
due to differences in our collection methodologies, we have found that the descriptions in the two overlapped datasets
exhibit significant qualitative differences, with descriptions in the UNC-Ref dataset tending to be more 
concise and to contain less flowery language than our descriptions.
\footnote{According to our personal communication with the authors of the UNC-Ref dataset, 
the instruction and reward rule of UNC-Ref
encourages the annotators to give a concise description in
a limited time, while in our G-Ref dataset, we encourage the annotators
to give rich and natural descriptions. This leads to
different styles of annotations.}
More specifically, the average lengths of expressions from our dataset and UNC-Ref are 8.43 and 3.61 respectively.
And the size of the word dictionaries (keeping only words appearing
more than 3 times) from our dataset and UNC-Ref are 4849 and
2890 respectively.
See 
Figure~\ref{fig:dataset_comp} for some visual comparisons.

\section{Tasks}
\label{sec:tasks}

In this section, we describe at a high level how we solve the two main
tasks of description and generation. We will describe the
model details and training in the next section.

\subsection{Generation}

In the \emph{description generation task}, 
the system is given a full image and a target object
(specified via a bounding box), and  it must generate a referring
expression for the target object.  
Formally, the task is to compute $\argmax_S p(S|R,I)$, where $S$ is a
sentence, $R$ is a region, and $I$ is an image.

Since we will use RNNs to represent $p(S|R,I)$,
we can generate $S$ one word at a time until we generate an end of
sentence symbol.
Computing the globally most probable sentence is hard,
but we can use beam search to approximately find the most probable sentences
(we use a beam size of 3). 
This is very similar to a standard image captioning task, except the
input is a region instead of a full image. The main difference is that
we will train our model to generate descriptions that
distinguish the input region from other candidate regions.

\subsection{Comprehension}

 In the
\emph{description comprehension task}, we are given a full image and a
referring expression and are asked to localize the 
the object being referred to  within the image by returning a bounding box.
One approach would be to train a model to directly predict the
bounding box location given the referring expression (and image).
However, in this paper, we adopt a simpler, ranking-based approach.
In particular, we first generate a set $\mathcal{C}$ of region
proposals, and then ask the system to rank these by probability.
Then we select the region using
$R^* = \argmax_{R \in \mathcal{C}} p(R|S,I)$,
where, by Bayes' rule, we have
\begin{equation}
p(R|S,I) = \frac{p(S|R,I) p(R|I)}{\sum_{R' \in \mathcal{C}} p(S|R',I) p(R'|I)}.
\label{eqn:postR}
\end{equation}
If we assume a uniform prior for
$p(R|I)$,\footnote{
This implies that we are equally
likely to choose any region to describe.
This is approximately true by virtue of the way we constructed the
dataset.  However, in real applications, region saliency $p(R|I)$ should be
taken into account.
} %
we can select the region
using $R^* = \argmax_{R \in \mathcal{C}} p(S|R,I)$.
This strategy is similar to  image retrieval methods
such as \cite{kiros2013multimodal,mao2014deep},
where the regions play the role of images.

At test time, we use the multibox method of \cite{erhan2014scalable}
to generate objects proposals. This generates a large number of class
agnostic bounding boxes.
We then classify each box into one of the 80 MS-COCO categories,
and
discard those with low scores. We use the resulting
post-classification boxes as the proposal set $\mathcal{C}$.
To get an upper bound on performance, we also use the ground truth
bounding boxes for all the objects in the image.
In both cases, we do not use the label for the object of
interest when ranking proposals.


\section{The Baseline Method}
\label{sec:baseline}

\begin{figure}[t] 
\begin{center}
\includegraphics[width=0.98\linewidth]{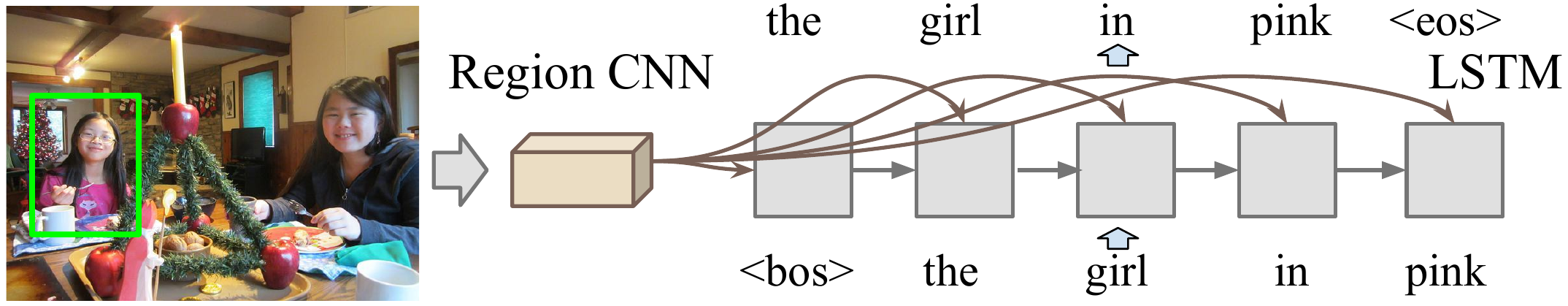}
\end{center}
\vspace{-0.3cm}
   \caption{llustration of the baseline model architecture.
$\langle$bos$\rangle$ and $\langle$eos$\rangle$ stand for beginning and end of sentence.
   		   }
\vspace{-0.5cm}
\label{fig:illu_model_baseline}
\end{figure}

In this section we explain our baseline method for
computing $p(S|R,I)$.

\subsection{Model Architecture}
\label{sec:baseline_model_arch}

Our baseline model is similar to other image captioning models that use a CNN
to represent the image, followed by an LSTM 
to generate the text (see
e.g., \cite{mao2014deep,donahue2014long,vinyals2014show}).  The main
difference is that we augment the CNN representation of the whole
image with a CNN representation of the region of interest, in addition to location information.
See Figure~\ref{fig:illu_model_baseline} for an illustration of our baseline
model.

In more detail, we use
VGGNet~\cite{simonyan2014very} as our CNN, pre-trained on the ImageNet
dataset \cite{deng2009imagenet,krizhevsky2012imagenet}.  The last 1000
dimensional layer of VGGNet is used as our representation of the
object region.
In addition, we compute features for the whole image, to serve as context.
In experiments, we only fine-tuned the weights for the last layer of the
CNN and fixed all other layers.  To feed a region to the CNN, we keep
the aspect ratio of the region fixed and scale it to $224 \times 224$
resolution, padding the margins with the mean pixel value (this is
similar to the region warping strategy in~\cite{girshick2014rcnn}).
This gives us a 2000-dimensional feature vector, for the region and image.

We encode the relative location and size of the region
using a 5 dimensional vector as follows: $[\frac{x_{tl}}{W},
\frac{y_{tl}}{H}, \frac{x_{br}}{W}, \frac{y_{br}}{H},
\frac{S_{bbox}}{S_{image}}]$, where $(x_{tl}, y_{tl})$ and $(x_{br},
y_{br})$ are the coordinates of the top left and bottom right corners
of the object bounding box, $H$ and $W$ are height and width of the
image, and $S_{bbox}$ and $S_{image}$ are the sizes of the bounding
box and image respectively.

Concatenating with the region, image, and location/size features, we obtain a 2005-dimensional 
vector which we feed as input into an LSTM sequence model, which parameterizes
the form of the distribution $p(S|R,I)$.
For our LSTMs, we use a 1024-dimensional word-embedding space,
and 1024-dimensional hidden state vector.
We adopt the most commonly used vanilla LSTM structure \cite{greff2015lstm} and
feed the  visual representation as input to the LSTM at each time step.

\subsection{Maximum Likelihood Training}
\label{sec:baseline_training}

 Our training data (discussed in
Section~\ref{sec:datasets}) consists of observed triplets $(I, R, S)$,
where $I$ is an image, $R$ denotes a region within $I$, and $S$ denotes a
referring expression for $R$.  To train the baseline model, we
minimize the negative log probability of the referring expressions
given their respective region and image:
\begin{equation}
\small
J(\theta) = -\sum_{n=1}^N \log p(S_n | R_n, I_n,\theta),
\label{eqn:ML}
\end{equation} 
where $\theta$ are the parameters of the RNN and CNN,
and where we sum over the $N$ examples in the training set.
We use ordinary stochastic gradient decent with a
batch size of 16 and use an initial learning rate of 0.01 which is
halved every 50,000 iterations.  Gradient norms are clipped to a
maximum value of 10.  To combat overfitting, we regularize using
dropout with a ratio of 0.5 for both the word-embedding and output
layers of the LSTM.

\section{The Full Method}
\label{sec:enhanced}

\begin{figure}[!tb]
\begin{center}
\includegraphics[width=0.98\linewidth]{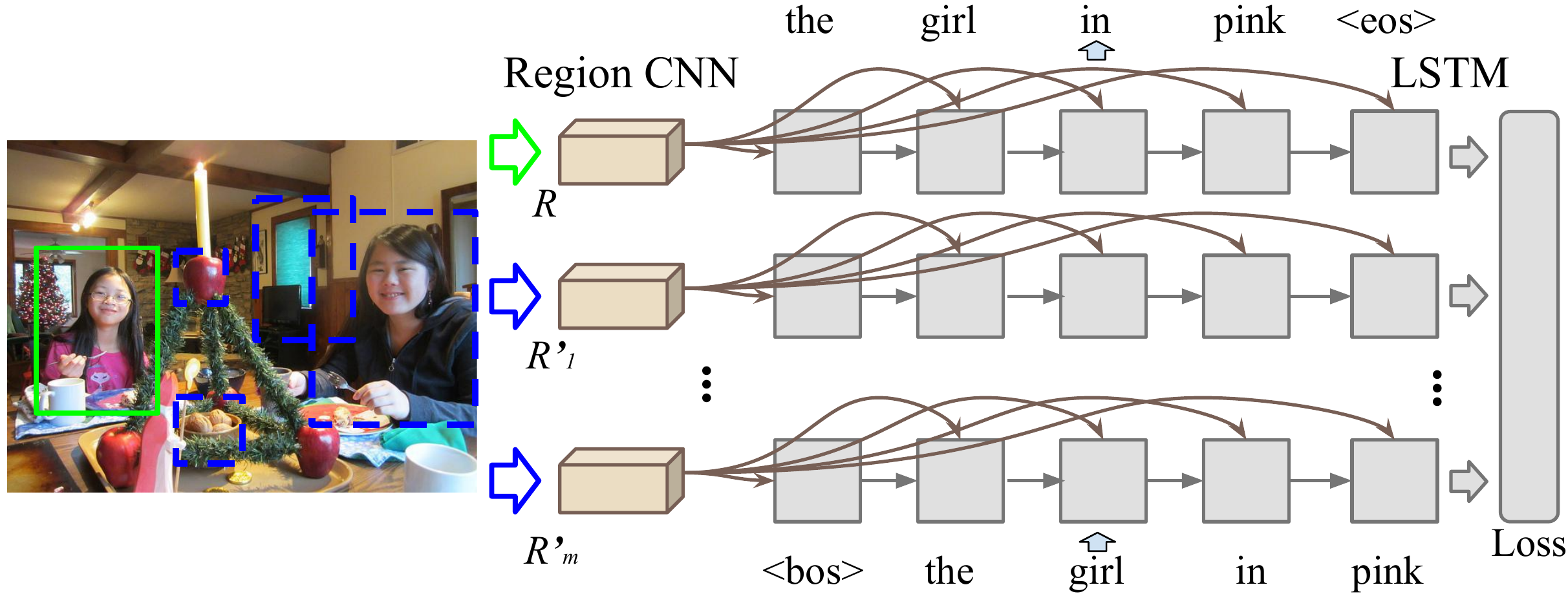}
\end{center}
\vspace{-0.3cm}
   \caption{Illustration of how we train the full model
using the softmax loss function.
$R$ (green) is the target region, $R'$ are the incorrect regions.
The weights of the LSTMs and CNNs are shared for $R$ and $R'$s.
(Best viewed in color)
\vspace{-0.5cm}
}
\label{fig:illu_model_softmax}
\end{figure}

The baseline method is to train the model to maximize $p(S|R,I)$, 
as is common for CNN-LSTM based image captioning models.
However a strategy that directly generates an expression based only on
the target object (which~\cite{golland2010game} calls the
\emph{reflex speaker} strategy) has the drawback that it may fail to
generate discriminative sentences.
For example, consider
Figure \ref{fig:illu_model_baseline}:
to generate a description of the girl highlighted
by the green bounding box,
generating the word ``pink'' is useful since it
distinguishes this girl from the other girl on the right. 
 To this end, we propose a modified training objective, described
 below.

\subsection{Discriminative (MMI) Training}
\label{sec:MMI}

Section~\ref{sec:baseline_training} proposed a way to train the model
using maximum likelihood. We now propose the following
alternative objective function:
\begin{equation} 
\small{
J'(\theta) = -\sum_{n=1}^N \log p(R_n | S_n, I_n,\theta),
}
\label{eqn:softmax}\vspace{-2mm}
\end{equation} 
where
\begin{equation} 
\small{
\log p(R_n | S_n, I_n,\theta) = \log 
\frac{p(S_n|R_n,I_n,\theta)}{\sum_{R' \in \mathcal{C}(I_n)}
  p(S_n|R',I_n,\theta)}.
}
\end{equation}
We will call this the softmax loss.
Note that this is the same as maximizing the mutual information
between $S$ and $R$ (assuming a uniform prior for $p(R)$), since
\begin{equation}
\small{
\mathrm{MI}(S, R) 
= \log \frac{p(S,R)}{p(R)p(S)}
= \log \frac{p(S|R)}{p(S)}.
}
\end{equation}
where $p(S) = \sum_{R'} p(S|R') p(R') = \sum_{R'} p(S|R')$.
Hence this approach is also called Maximum Mutual Information (MMI) training
\cite{Bahl1986}.

The main intuition behind MMI training is that we want to consider whether a listener
would interpret the sentence unambiguously.  We do this by penalizing the model if it thinks that a 
referring expression for a target object could also be plausibly generated by 
some other object within the same image.
Thus given a training sample $(I, R, S)$, we train a model that outputs a high 
$p(S\,|\,R,I)$, while maintaining a low $p(S\,|\,R',I)$, whenever $R'\neq R$.
Note that this stands in contrast to the Maximum Likelihood (ML) objective function in Equation \ref{eqn:ML} which
directly maximizes $p(S|R)$ without considering other objects in the image. 



There are several ways to select the region proposals $\mathcal{C}$.
We could use all the true object bounding boxes, but this tends to
waste time on objects that are visually very easy to discriminate from
the target object (hence we call these \textit{``easy ground truth
negatives''}).
An alternative is to select true object bounding boxes belonging to
objects of the same class as the target object; these are more
confusable (hence we call them \textit{``hard ground truth negatives''}).
Finally, we can use multibox proposals, the same as we use at test
time, and select the ones with the same predicted object labels as $R$
(hence we call them \textit{``hard multibox negatives''}).
We will compare these different methods in Section~\ref{sec:exp_diag}.
We use 5 random negatives at each step, so that all the data for a
given image fits into GPU memory.

To optimize Equation~\ref{eqn:softmax}, we must replicate the network
(using tied weights) for each  region $R' \in \mathcal{C}(I_n)$ (including
the true region $R_n$), as shown in Figure~\ref{fig:illu_model_softmax}.
The resulting MMI trained model has exactly the same number of parameters as the ML trained model, and 
we use the same optimization and regularization strategy as in
Section~\ref{sec:baseline_training}.  Thus the only difference is the
objective function.

For computational reasons, it is  more convenient to use 
the following max-margin loss, which compares the target region $R$
against a single random negative region $R'$:
\begin{equation}\footnotesize
\begin{split}
J''(\theta) = - & \sum_{n=1}^N \{\log p(S_n|R_n, I_n, \theta) -\\
\lambda \max( 0,   
& \;\;  M - \log p(S_n | R_n, I_n, \theta) + \log p(S_n|R_n', I_n, \theta) ) \}
\end{split}
\label{eqn:maxmargin}
\end{equation}
This objective, which we call max-margin MMI (or MMI-MM) intuitively captures a
similar effect as its softmax counterpart (MMI-SoftMax)
and as we show in Section~\ref{sec:exp_diag}, yields similar
results in practice.
However, since the max-margin objective only compares two regions, the network must only be replicated twice.
Consequently, less memory is used per sentence, allowing for more sentences to
be loaded per minibatch which in turn helps in stabilizing the gradient.


\vspace{-0.2cm}
\section{Semi-supervised Training}
\label{sec:semisuper}

\begin{figure}[t!]
\begin{center}
\includegraphics[width=0.98\linewidth]{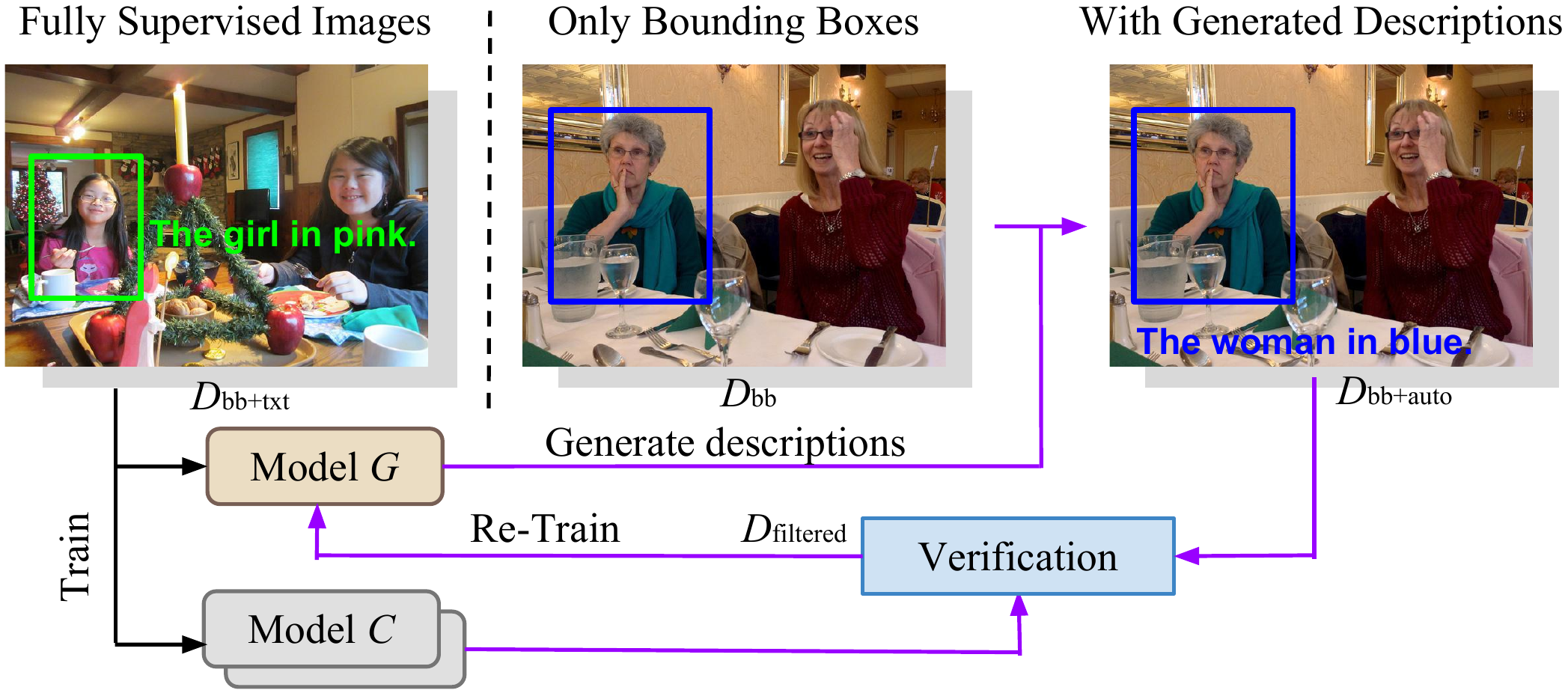}
\end{center}
\vspace{-0.3cm}
   \caption{Ilustration of the semi-supervised training process.
See text for details.
   		   }
\vspace{-0.5cm}
\label{fig:illu_semi_supervised}
\end{figure}

\newcommand{\Dfull}{D_{\mathrm{bb+txt}}}
\newcommand{\Dpartial}{D_{\mathrm{bb}}}
\newcommand{\Dnew}{D_{\mathrm{bb+auto}}}
\newcommand{\Dfilt}{D_{\mathrm{filtered}}}

Collecting referring expressions data can be expensive.
In this section we discuss semi-supervised training of our full model by making use of
bounding boxes that do not have descriptions, and thus are more ubiquitously available.
Our main intuition for why a bounding box (region) $R$ can be useful even without an accompanying description 
is because it allows us to penalize our model during MMI training if it generates a sentence  that it cannot itself decode to correctly recover $R$ 
(recall that MMI encourages $p(S | R, I)$ to be higher than $p(S | R', I)$, whenever $R' \neq R$).

In this semi-supervised setting, we
consider a small dataset $\Dfull$ of images with bounding boxes and
descriptions, together with a larger dataset $\Dpartial$ of images and  bounding
boxes, but without descriptions. We use $\Dfull$ to train a model
(which we call model $G$)  
to compute $p(S|R,I)$. We then use this model $G$ to generate a set of
descriptions for the bounding boxes in $\Dpartial$ (we call this new dataset $\Dnew$).
We then retrain $G$ on $\Dfull \cup \Dnew$, in the spirit of bootstrap
learning.

The above strategy suffers from the flaw
that not all of the generated sentences are reliable, which may ``pollute'' the training set.
To handle this, we train an ensemble of different models
on $\Dfull$ (call them model $C$),
and use these to determine which of the generated sentences for $\Dnew$
are
trustworthy. In particular, we apply each model in the ensemble to decode each
sentence in $\Dnew$, and only keep the sentence if every model 
maps it to the same correct object; we will call the resulting
verified dataset $\Dfilt$.
This ensures that the generator creates referring expressions that can
be understood by a variety of different models, thus minimizing overfitting.
See Figure~\ref{fig:illu_semi_supervised} for an illustration.
In the experiments, we show that our model benefits from this semi-supervised training.
\eat{\KEVIN{Do you show that filtering is necessary/ useful compared to
  naive bootstrapping?}}

\eat{
We are curious about how to use less data to train a good model since the fully annotation of an texture object description dataset is expensive.
On the contrast, it is easy to get lots of images with object bounding box annotation. 
To utilize these data, we proposed a bootstrapping method to improve our model under a semi-supervised setting, as illustrated in Figure \ref{fig:illu_semi_supervised}.
More specifically, we firstly train a model using the images with object descriptions.
We then use this model to generate object descriptions for the images with only object bounding boxes.
The simplest way is to directly add these descriptions to the fully supervised training set and finetune the model.
However, the generated sentences often contains lots of noisy and errors that might be harmful for the training.

Our solution is that in addition to the initial model trained to generate these descriptions (denoted as Model(G)), we train several models at the same time with different initialization of the parameters and different data feeding order.
These models (denoted as Models(C)) will be used to verified the generated descriptions by Model(G)in the object description comprehension task.
A generated description can pass the verification step only if all of Models(C) successfully predict the object bounding box according to the description.
In the experiments, we show that our model benefits from this semi-supervised training.
}

\section{Experiments}

We conducted experiments on both of the COCO referring expression
datasets mentioned in Section~\ref{sec:datasets}:
our G-Ref dataset and the UNC-Ref dataset.
We randomly chose 5,000 objects as the validation set, 5,000 objects
as the testing set and the remaining objects  as the
training set (44,822 for G-Ref and 40,000 for UNC-Ref).
\eat{
\KEVIN{Wouldn't it be better to partition at the image level?}
\Junhua{Yes. But I somehow use this partition at the beginning and have done most of the experiments when I realize it. We may need to change the strategy when I want to release the dataset publicly. However, I think the current setting does not decrease the difficulties of the task.}
}

\subsection{Evaluation Metrics} 
\label{sec:exp_eval}

In this section, we describe how we evaluate performance of the comprehension and generation tasks.

The comprehension task is easy to evaluate: we simply
compute the Intersection over Union (IoU)
ratio between the true and predicted bounding box.
If IoU exceeds 0.5, we call the
detection a true positive, otherwise it is a false positive (this is
equivalent to computing the precision$@1$ measure).
We then average this score over all  images.

The generation task is more difficult ---
we can evaluate a generated description in the same way as an image description, using metrics such as 
CIDEr \cite{Vedantam2015cider},
BLEU \cite{papineni2002bleu} and METEOR \cite{lavie2007meteor}.
However these metrics can be unreliable and do not account for semantic meaning.
We rely instead on human evaluation, as was done in the most recent
image captioning competition \cite{captioncoco}.
In particular, we asked Amazon Mechanical Turk (AMT) workers
to compare an automatically generated object description to a human
generated object description, when presented with an image and object
of interest.
The AMT workers do not
know which sentences are human generated and which are
computer generated (we do not even tell them that some
sentences might be computer generated to reduce possible bias).
We simply ask
them to judge which sentence is a better description, or if
they are equally good.

In addition to human evaluation, which does not scale, we evaluate our entire system
by passing automatically generated descriptions to our comprehension system, and verifying that 
they get correctly decoded to the original object of interest.  This end-to-end test
is automatic and much more reliable than standard image captioning metrics.
%

\subsection{Comparing different training methods}
\label{sec:exp_diag}

\begin{table}[!tb] \small
\begin{center}
\begin{tabular}{lcccc} \hline \multicolumn{1}{r|}{Proposals} &
\multicolumn{2}{c|}{GT} & \multicolumn{2}{c}{Multibox} \\
\multicolumn{1}{r|}{Descriptions} & GEN & \multicolumn{1}{c|}{GT} &
GEN & GT \\ \hline 
ML (baseline) & 0.803 & 0.654 & 0.564 & 0.478 \\
MMI-MM-easy-GT-neg & 0.851 & 0.677 & 0.590 & 0.492 \\
MMI-MM-hard-GT-neg & \textbf{0.857} & \textbf{0.699} & 0.591 & 0.503\\
MMI-MM-multibox-neg & 0.848 & 0.695 & \textbf{0.604} &
\textbf{0.511} \\
\hdashline
MMI-SoftMax & 0.848 & 0.689 & 0.591 & 0.502 \\
\hline
\end{tabular}%
\end{center}
	\caption{We measure precision$@1$ on the UNC-Ref validation
          data. Each row is a different way of training the model. The
          columns show performance on ground truth or multibox
          proposals, and ground truth (human) or generated
          descriptions.
Thus the columns with GT descriptions evaluate the performance of the
comprehension system, and the columns with GEN descriptions evaluate
(in an end-to-end way) the performance of the generation system.
}
\vspace{-0.3cm}
	\label{tab:diag}
\end{table}

In this section, we compare different ways of training our model:
maximum likelihood training (the baseline method);
max-margin loss with easy ground truth negatives (``MMI-MM-easy-GT-neg'');
max-margin loss with hard ground truth negatives (``MMI-MM-hard-GT-neg'');
max-margin loss with hard multibox negatives (``MMI-MM-multibox-neg'');
softmax/MMI loss with hard multibox negatives (``MMI-SoftMax'').
For each method, we consider using either ground truth or multibox
proposals at test time. In addition, we consider both ground truth
descriptions and generated descriptions. 

In this experiment we treat UNC-Ref as a validation
set to explore various algorithmic options and hyperparameter
settings for MMI. 
Only after having fixed these algorithmic options
and hyperparameter settings did we do experiments
on our G-Ref dataset (Section~\ref{sec:exp_fully_train}).
This reduces the risk that we will have ``overfit'' our hyperparameters to each particular
dataset.
The results are summarized in Table \ref{tab:diag} and we draw the following conclusions:
\begin{itemize}[noitemsep,topsep=0pt,parsep=0pt,partopsep=0pt]
\item All 
models perform better on generated descriptions than the groundtruth
ones, possibly because the generated descriptions are shorter than the
groundtruth (5.99 words on average vs 8.43),
and/or because the generation and comprehension
models share the same parameters, so that even if the generator uses a
word incorrectly (e.g., describing a ``dog'' as a ``cat''), the
comprehension system can still decode it correctly.
Intuitively, a model might ``communicate'' better with itself using its own language than with others.

\item All the variants of the Full model (using MMI training) work
better than the strong baseline using maximum likelihood training.

\item The softmax version of MMI training is similar to the max-margin
  method, but slightly worse.

\item MMI training benefits more from hard negatives than easy ones.

\item Training on ground truth negatives helps when using
ground truth proposals, but when using multibox proposals (which is
what we can use in practice), it is
better to use multibox negatives.
\end{itemize}

Based on the above results, for the rest of the paper
we will use max-margin training with hard multibox negatives as our Full Model.

\subsection{Fully-supervised Training}
\label{sec:exp_fully_train}

\begin{table}[!tb] \small
\begin{center}
\begin{tabular}{lcccc} 
\hline 
\multicolumn{1}{r|}{Proposals} & \multicolumn{2}{c}{GT} & \multicolumn{2}{c}{multibox} \\
\multicolumn{1}{r|}{Descriptions} & GEN & GT & GEN & GT \\
\hline 
\multicolumn{5}{c}{G-Ref-Val} \\
\hline
Baseline & 0.751 & 0.579 & 0.468 & 0.425 \\ 
Full Model & \textbf{0.799} & \textbf{0.607} & \textbf{0.500} & \textbf{0.445} \\
\hline 
\multicolumn{5}{c}{G-Ref-Test} \\
\hline 
Baseline & 0.769 & 0.545 & 0.485 & 0.406 \\
Full Model & \textbf{0.811} & \textbf{0.606} & \textbf{0.513} & \textbf{0.446} \\
\hline
\multicolumn{5}{c}{UNC-Ref-Val} \\
\hline
Baseline & 0.803 & 0.654 & 0.564 & 0.478 \\
Full Model & \textbf{0.848} & \textbf{0.695} & \textbf{0.604} & \textbf{0.511} \\
\hline
\multicolumn{5}{c}{UNC-Ref-Test} \\
\hline
Baseline & 0.834 & 0.643 & 0.596 & 0.477 \\
Full Model & \textbf{0.851} & \textbf{0.700} & \textbf{0.603} & \textbf{0.518} \\
\hline
\end{tabular}%
\end{center}
\vspace{-0.2cm}
	\caption{Precision$@1$ for the baseline (ML) method and our full model with the max-margin
          objective function on various datasets.
}
\vspace{-0.3cm}
	\label{tab:fully_results}
\end{table}

In this section, we compare the strong baseline (maximum likelihood)
with our max-margin MMI method on the validation and test sets from
G-Ref and UNC-Ref. As before, we consider ground truth
and multibox proposals at test time, and ground truth (human) or
generated (automatic) descriptions.
The results are shown in Table \ref{tab:fully_results}.  
We see that MMI training outperforms ML training under every setting.
\footnote{We also train our baseline and full model on a random train, val, and test split
w.r.t. to the images of our G-Ref dataset.
The results are consistent with those in Table \ref{tab:fully_results}.
With multibox proposals and GT descriptions, the Precision$@1$ of the baseline and full model are
0.404 and 0.444 on val set, and 0.407 and 0.451 on test set respectively.
}

In addition to the above end-to-end evaluation,
we use human evaluators to judge generated sentence quality.
In particular, we selected 1000 objects at random from our test set,
and showed them to Amazon Mechanical Turk workers. The percentage of descriptions that are
evaluated as better or equal to a human caption for the baseline and the
full model are 15.9\% and 20.4\% respectively.  This shows that MMI
training is much better (4.5\% absolute improvement, and 28.5\%
relative) than ML training.

\eat{
\KEVIN{I dropped the paragraph on BLEU, since we have already
  criticized that metric.}
We calculate the BLEU-4 score \cite{papineni2002bleu} for the  baseline and our full model.
The scores are 0.129 and 0.123 respectively.
Since BLEU emphasizes the smoothness of the sentence rather than the semantic information, 
the image caption baseline performs slightly better than our Full model.
But in our task, we cares more about the discriminative ability of methods, which is essential for an effective communication.
BLEU fails to capture such information. 
\eat{To put this number in perspective, in the recent image captioning
contest for MSCOCO, the best performing system (they fintuned the CNN in their model while we do not) generated captions that
were as good or better than human captions 27.3\% of the time \cite{captioncocoresults}.
\Junhua{Do we need to mentioned the results of MSCOCO here? The task is different.}
\Jon{mention that they finetune the CNN, which we don't do?}}
}

\subsection{Semi-supervised Training}
\label{sec:exp_semi_train}

\eat{
To conduct the semi-supervised training experiment, we separate the
training set of our G-Ref dataset and the UNC-Ref dataset into two parts with the
same number of objects.  The first part (denoted by $\Dfull$)
has the object description annotations while the second part (denoted
by $\Dpartial$) only has object bounding boxes.  
As explained in Section~\ref{sec:semisuper}, we first train a model on $\Dfull$
and use this to generate descriptions for the objects in $\Dpartial$;
we then filter the generated expressions
using an ensemble of 2 other models (trained on $\Dfull$)
to give the final $\Dfilt$. Finally we retrain on $\Dfull \cup
\Dfilt$.
}

To conduct the semi-supervised training experiment, we separate the
training set of our G-Ref dataset and the UNC-Ref dataset into two parts with the
same number of objects.  The first part (denoted by $\Dfull$)
has the object description annotations while the second part (denoted
by $\Dpartial$) only has object bounding boxes.  
Table \ref{tab:semi_results} shows the results of semi-supervised
training on the validation set of our dataset and UNC-Ref. We see that we get some
improvement by training on $\Dfull \cup \Dpartial$ over just using
$\Dfull$.

\begin{table}[!tb] \small
\begin{center}
\begin{tabular}{lcccc} 
\hline 
\multicolumn{1}{r|}{Proposals} & \multicolumn{2}{c}{GT} & \multicolumn{2}{c}{multibox} \\
\multicolumn{1}{r|}{Descriptions} & GEN & GT & GEN & GT \\
\hline 
\multicolumn{5}{c}{G-Ref} \\
\hline
$\Dfull$ & 0.791 & 0.561 & \textbf{0.489} & 0.417 \\
$\Dfull \cup \Dpartial$ & \textbf{0.793} & \textbf{0.577} & \textbf{0.489} & \textbf{0.424} \\
\hline 
\multicolumn{5}{c}{UNC-Ref} \\
\hline 
$\Dfull$ & 0.826 & 0.655 & 0.588 & 0.483  \\
$\Dfull \cup \Dpartial$ & \textbf{0.833} & \textbf{0.660} & \textbf{0.591} & \textbf{0.486} \\
\hline
\end{tabular}%
\end{center}
\vspace{-0.2cm}
	\caption{Performance of our full model when trained on a small strongly
          labeled dataset vs training on a larger dataset with automatically
          labeled data.
}
	\label{tab:semi_results}
\vspace{-0.3cm}
\end{table}

\subsection{Qualitative Results}

\begin{figure*}[thb] 
\begin{center}
\includegraphics[width=0.98\linewidth]{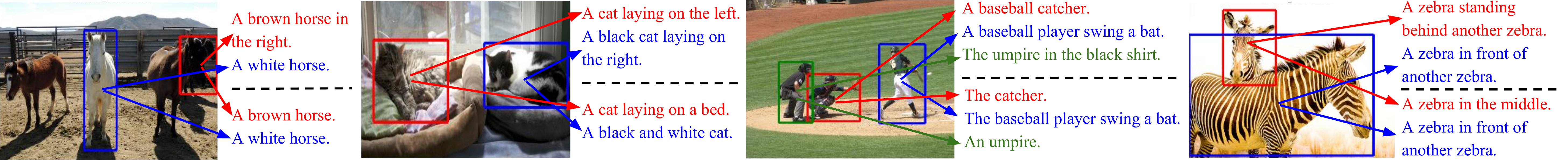}
\end{center}
\vspace{-0.3cm}
   \caption{The sample results of the description generation using our full model (above the dashed line) and the strong baseline (below the dashed line).
   			The descriptions generated by our full model are more discriminative than those generated by the baseline.
   		   }
\label{fig:speaker_results}
\end{figure*}

\begin{figure*}[thb] 
\begin{center}
\includegraphics[width=0.95\linewidth]{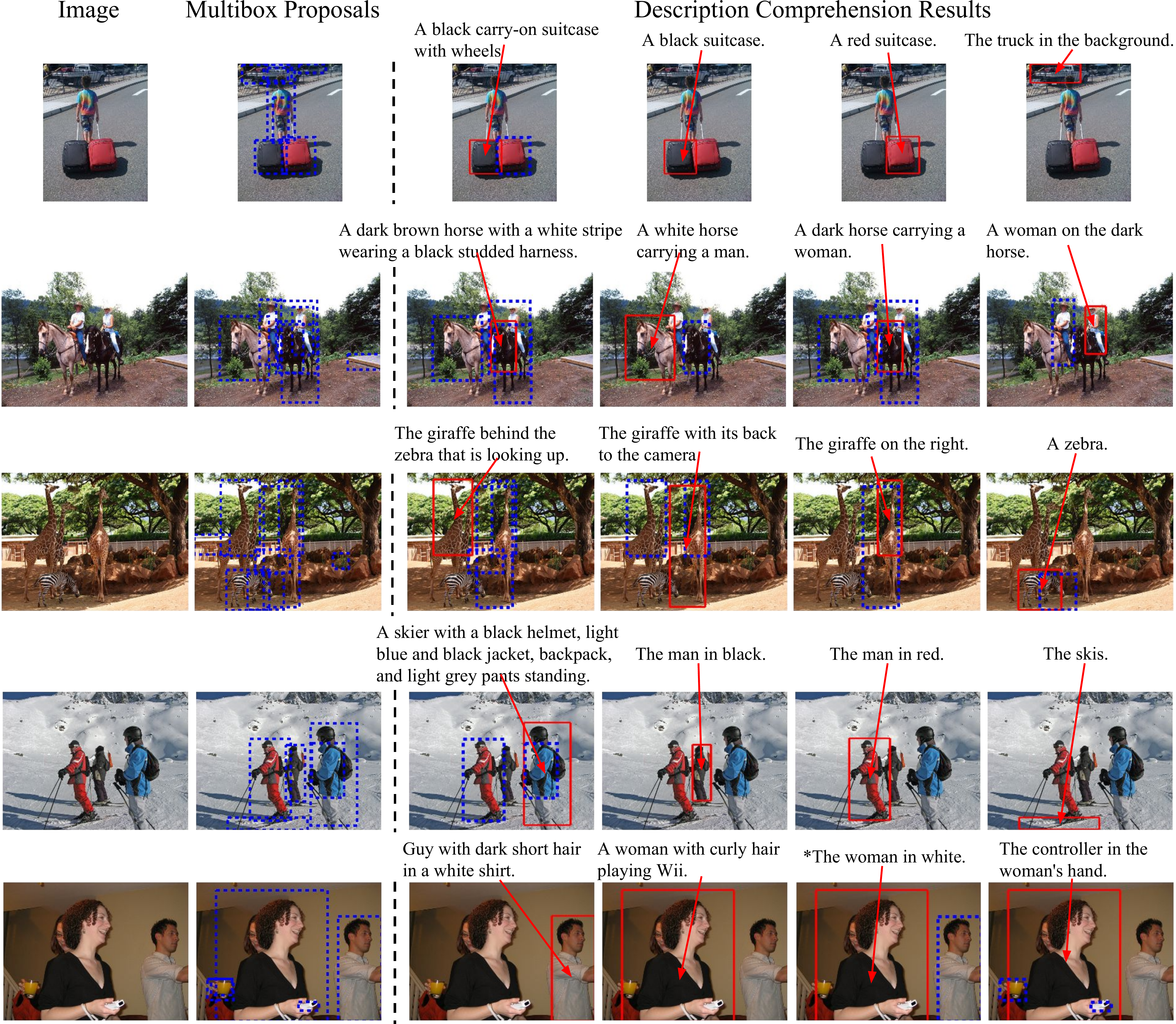}
\end{center}
\vspace{-0.3cm}
   \caption{Sample results of the description comprehension task using our full model.
   			The first and second column shows the original image and the multibox proposals.
   			The third to sixth columns show the results of our model when input an arbitrary description of an object in the image.
   			The red bounding box denotes the most probable object predicted by the model while the blue dashed ones denote the bounding boxes within the margin of the most probable one.
   			The descriptions can be the groundtruth ones in the dataset (third column) or an customized descriptions (fourth to sixth columns).
   			(Best viewed in color)
   		   }
\vspace{-0.3cm}
\label{fig:listener_results}
\end{figure*}

In Figure \ref{fig:speaker_results} 
we show qualitative results of our full generation model
(above the dashed line) and the baseline generation model (below the
dashed line) on some of our test images.
We see that the descriptions generated by our full model are typically longer and more
discriminative than the baseline model.
In the second image, for example, the baseline describes one of the cats as ``a cat laying on a bed'',
which is not sufficiently unambiguous for a listener to understand which cat is being described.
Our full model, on the other hand, describes the same cat as ``a cat laying on the left'' which is completely unambiguous.

Figure \ref{fig:listener_results} shows some qualitative results of
our full comprehension model on our test dataset.
The first and second columns show the original image and the multibox proposals respectively.
The last four columns show the bounding boxes (denoted as a red
bounding box in the figure) selected by our full model
in response to different input sentences (both ground truth sentences
and ones we created to probe the comprehension abilities of the model).
To better interpret these results, we also show the bounding boxes that are within the margin of the model (see Eqn. \ref{eqn:maxmargin}) with dashed blue bounding boxes.
Their bounding boxes are considered as ``possible candidates'' but
their scores (i.e. $p(S|R, I)$) are not as high as the chosen one. 

In general, we see that the comprehension model does quite well from short two word 
phrases to longer descriptions.  It is able to respond correctly to single word changes in a referring expression 
(e.g., ``the man in black'' to ``the man in red''). It also correctly identifies that the horse is the referent of the expression ``a dark horse
carrying a woman'' whereas the woman is the referent in ``a woman on the dark horse'' --- note that methods that average word
embeddings would most likely fail on this example.
However, there are also failure cases.
For example, in the fifth row, ``the woman in white'' selects a woman in
black; this is because our model cannot handle the case where the
object is not present, although it makes a reasonable guess.
Also, in the fifth row, ``the controller in the woman's hand'' selects the woman, the orange juice and the controller, since this
particular kind of object is too small to detect, and lacks enough training data.

\eat{
Since this task can be treated as an advanced object detection task with more freedom of input, we also try to input interesting customized descriptions to the model.
E.g., we test descriptions of some easy neglected objects (see the sixth images in the first and fourth row).
Our model successfully handles this challenge.
We also try descriptions that actually are consistent with no object in the images (see the fifth images in the fourth row).
The model makes a reasonable guess of the description.
Another interesting examples are the images in the fifth and sixth column in the second row.
To successfully predict the correct object, the model need to understand the importance of the order of the words in a sentence.
We show a failure case in the fifth row and sixth column.
The object ``ski board'' is often small in the image, and lacks enough training samples.
The model thinks almost all the proposals are possible candidates for the description.
Another common failure cases are due to the absence of the multibox
proposals that overlapped with the correct object.
}

\vspace{-0.1cm}
\section{Conclusions}\label{sec:conclusions}
\vspace{-0.1cm}
To conclude, we leave the reader with two simple points.
First, referring expressions have been studied for decades, but
in light of the recent burst of interest in image captioning, referring expressions
take on new importance.  Where image captioning itself is difficult to evaluate,
referring expressions have an objective performance metric,
and require the same semantic understanding of language and vision. 
 Thus success on datasets such as the one contributed in this paper is 
 more meaningful than success by standard image captioning metrics.

Second, to be successful at generating descriptions, we must 
\emph{consider the listener}.  Our experiments show that modeling 
a listener that must correctly decode a generated description consistently outperforms a model 
 that simply emits captions based on region features.  We hope that in addition to 
our dataset, these insights will spur further progress on joint models of vision and language.

\eat{
In this paper, we present a method that deals with the task of discriminative object description generation and comprehension jointly.
We construct a new large-scale object description dataset based on MS COCO dataset to validate our method, which will be released.
Compared to a strong baseline adapted from RNN-CNN based image captioning models, our method performs consistently better with various evaluation metrics.
We also present a semi-supervised training algorithm that allows us to bootstrap the model with unlabeled data.
In the future work, we will continue to explore the semi-supervised training task using large amount of unlabeled data.
}

\section*{Acknowledgement}

We are grateful to Tamara Berg for sharing the UNC-Ref-COCO dataset.
We also thank Sergio Guadarrama, Vivek Rathod, Vignesh Ramanathan, Nando de Freitas, Rahul Sukthankar, Oriol Vinyals and Samy Bengio for early discussions and feedback on drafts.
This work was partly supported by ARO 62250-CS, the NSF Center for Brains, Minds, and Machines, and NSF STC award CCF-1231216.

{\small
\bibliographystyle{ieee}
\bibliography{egbib}
}

\end{document}